\documentclass[letterpaper, 10 pt, conference]{ieeeconf}  

\usepackage{url}
\usepackage{graphicx}
\usepackage{amsmath,amsfonts}
\usepackage{algorithm}

\usepackage{algpseudocode}

\usepackage{array}
\usepackage[caption=false,font=normalsize,labelfont=sf,textfont=sf]{subfig}
\usepackage{textcomp}
\usepackage{stfloats}
\usepackage{hyperref}
\usepackage{verbatim}
\usepackage{cite}

\usepackage{amssymb}
\usepackage{color}
\usepackage{cleveref}

\usepackage{hyperref}
\usepackage{mathrsfs}
\usepackage{booktabs}
\usepackage{makecell}
\usepackage{multirow}
\usepackage{pifont}

\IEEEoverridecommandlockouts                              

\title{\LARGE \bf
TransforMARS: Fault-Tolerant Self-Reconfiguration for Arbitrarily Shaped Modular Aerial Robot Systems 
}

\author{Rui Huang, Zhiyu Gao, Siyu Tang, Jialin Zhang, Lei He, Ziqian Zhang, Lin Zhao
\thanks{Rui Huang,  Siyu Tang, Jialin Zhang, Lei He, Ziqian Zhang and Lin Zhao are with the Department of Electrical and Computer Engineering, National University of Singapore, Singapore 117583 (Email: {ruihuang@u.nus.edu, e1352616@u.nus.edu, jialinzhang@u.nus.edu, lei.he@nus.edu.sg, e1546905@u.nus.edu,  zhaolin@nus.edu.sg}). Zhiyu Gao is with the Department of Mechanical  Engineering, National University of Singapore, Singapore 117583 (Email: {e1553948@u.nus.edu}).
This work was supported by the Singapore Ministry of Education Tier 2 Academic Research Fund (T2EP20123-0037, T2EP20224-0035). Correspondence to: Lin Zhao.}
}
\begin{document}

\maketitle
\thispagestyle{empty}
\pagestyle{empty}

\begin{abstract}
Modular Aerial Robot Systems (MARS) consist of multiple drone modules that are physically bound together to form a single structure for flight. Exploiting structural redundancy, MARS can be reconfigured into different formations to mitigate unit or rotor failures and maintain stable flight. Prior work on MARS self-reconfiguration has solely focused on maximizing controllability margins to tolerate a single rotor or unit fault for rectangular-shaped MARS. We propose TransforMARS, a general fault-tolerant reconfiguration framework that transforms arbitrarily shaped MARS under multiple rotor and unit faults while ensuring continuous in-air stability.
Specifically, we develop algorithms to first identify and construct minimum controllable assemblies containing faulty units. We then plan feasible disassembly-assembly sequences to transport MARS units or subassemblies to form target configuration. Our approach enables more flexible and practical feasible reconfiguration. We validate TransforMARS in challenging arbitrarily shaped MARS configurations, demonstrating substantial improvements over prior works in both the capacity of handling diverse  configurations and the number of faults tolerated. The videos and source code of this work are available at \url{https://github.com/RuiHuangNUS/TransforMARS}
\end{abstract}


\section{Introduction}

Modular Aerial Robotic Systems (MARS) serve as flexible and adaptive agents in diverse flight tasks including searching, rescuing, and transportation. MARS can respond to dynamical environment changes through disassembly and reassembly. Research on MARS has developed mid-air docking~\cite{Modquad, ModQuad-Vi, TokyoU-sugihara2023design}, in-flight separation~\cite{separation-2019design, zhang2024design, TokyoU-sugihara2024beatle, AirTwins2023}, self-reconfiguration algorithms~\cite{saldana2017decentralized, SR-baseline2020, SR-ICRA2025}, structural optimization~\cite{gabrich2021finding, Jiawei2023ICRA, su2024flight,gu2025modur}, trajectory tracking ~\cite{H-modquad, Jiawei2025}, fault-tolerance control ~\cite{Mengguang2024ICRA, huang2025robust-iros} and collision-free path planning~\cite{huang2025robust-iros,gu2026self} to improve robustness and flexibility. However, compared with ground modular robots~\cite{ground-luo2022auto, ground-zhao2025modular, ground-10802149, ground-sun2023mean}, aerial platforms are more severely affected by faults, which can severely degrade control performance if not properly mitigated. 

\begin{figure}[!t]
\centering

\includegraphics[width=8.5cm]{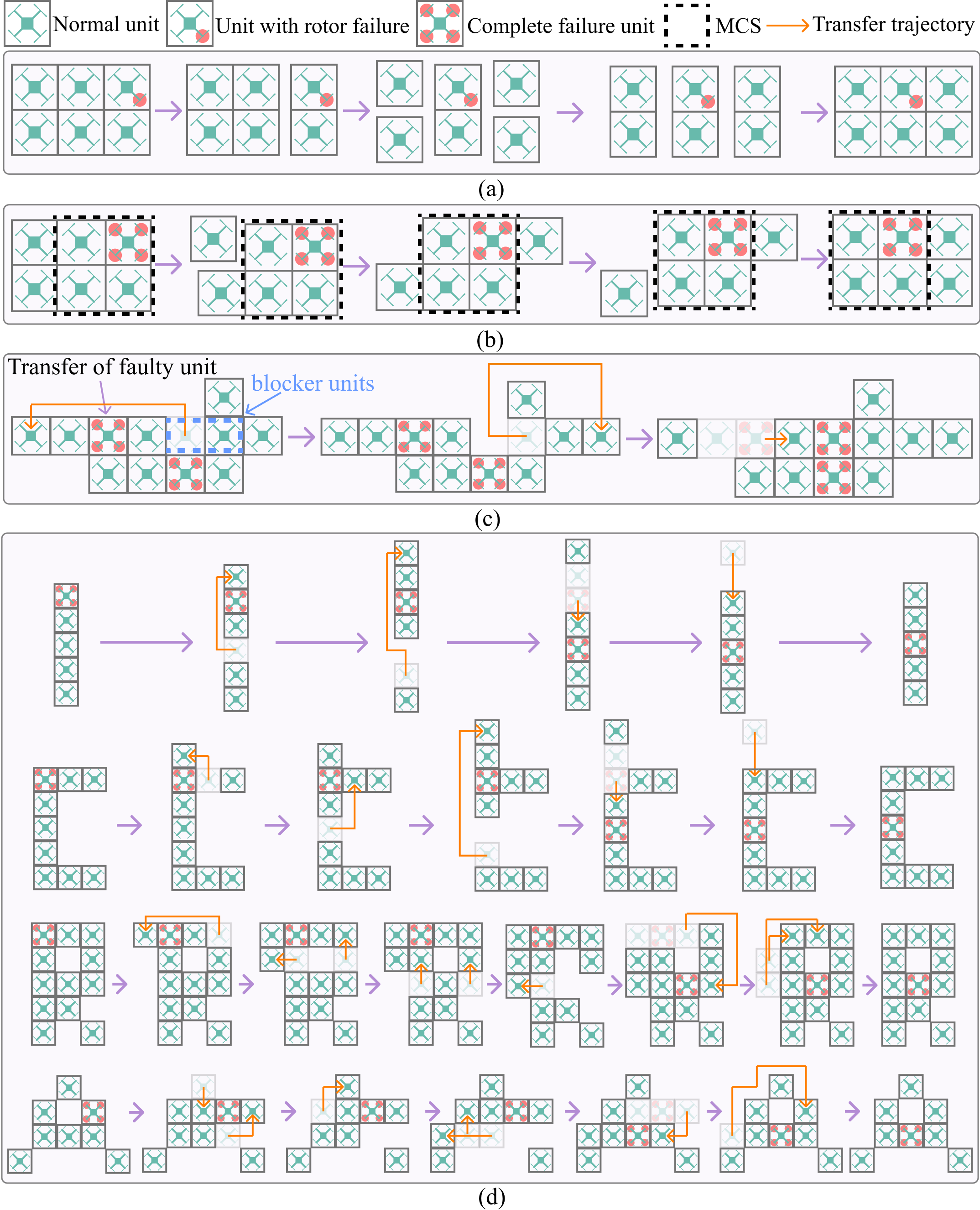}
\vspace{-3mm}
\caption{Self-reconfiguration of $3\times2$ and arbitrary assemblies. (a) Prior work~\cite{SR-baseline2020} attaches a functional unit adjacent to a faulty one, addressing only rotor-level failure. (b) Subsequent work~\cite{SR-ICRA2025} optimized controllability and extended the approach to single-unit faults, but remained limited to rectangular configurations.
 (c) Proposed TransforMARS accommodates multiple unit faults in arbitrary configurations; in the first two steps, normal units are relocated to create vacancies for the minimum controllable subassembly containing the faulty units. (d) TransforMARS reconfigures in ICRA letter-shaped formations, demonstrating flexibility under faulty units.
}
\label{fig:demo}
\vspace{-8mm}
\end{figure}

To address this challenge, prior studies~\cite{SR-baseline2020, SR-ICRA2025} have shown that self-reconfiguration can significantly improve trajectory-tracking performance after faults occur. Specifically, prior work~\cite{SR-baseline2020} proposed a self-reconfiguration strategy to enhance fault-tolerant control under rotor-level failures, thereby maintaining mission continuity. The approach formulates a mixed-integer linear program that selects module placements to mitigate fault effects. However, the scheme has clear limitations. As illustrated in Fig.\ref{fig:demo}(a), it primarily targets partial faults (e.g., a single failed rotor in a quadcopter) by attaching a functional module to the impaired unit (Step 2), and it does not handle complete unit failure. Moreover, it does not explicitly account for uncontrollable factors, and its performance depends on both the fidelity of the simulator’s physics engine and the selected objective function for arranging the position of the faulty unit. Subsequent work\cite{SR-ICRA2025} narrows this gap by deriving a control-constrained dynamic model of MARS and proposing a robust self-reconfiguration algorithm that maximizes the Controllability Margin (CM) at each intermediate reconfiguration step. In particular, it computes a Minimally Controllable Subassembly (MCS) that encompasses the faulty unit to enable stable in-air transport, and then plans control-feasible disassembly and assembly sequences for robust self-reconfiguration, illustrated in Fig.~\ref{fig:demo}(b).

Nevertheless,~\cite{SR-ICRA2025} still exhibits two key limitations: (i) \textit{limited generalization}--it assumes only a single faulty unit and confines the analysis to standard rectangular configurations, so it cannot scale to multiple faults or arbitrary configurations; and (ii) \textit{lack of path planning}--it omits explicit collision-aware motion planning and conflict-free assembly during self-reconfiguration. Consequently, the routing of units fails to generalize across configurations.

\begin{figure*}[!t]
\centering
\includegraphics[width=17.5 cm]{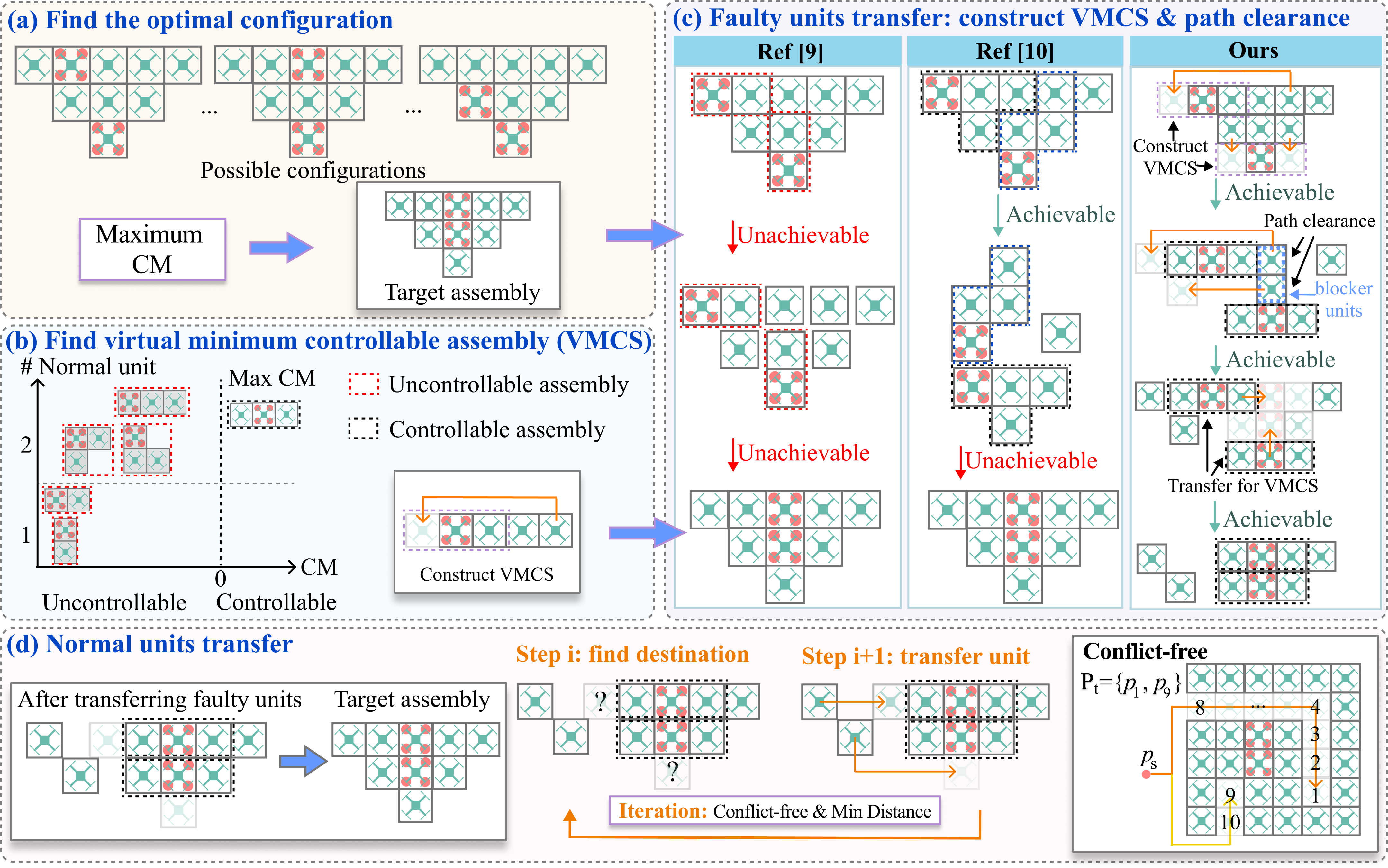}
\vspace{-4 mm}
\caption{Framework of TransforMARS. (a) The configuration with the maximum CM value is computed as the target assembly. (b) At each step, a normal unit is connected to the faulty unit until the CM becomes positive, and the VMCS is chosen as the configuration with the maximum CM under the same normal unit. (c) The VMCS is then constructed by relocating suitable normal units, followed by path clearance to ensure all VMCS units can reach their designated positions. (d) The remaining vacant positions are sequentially filled by selecting non-conflicting destinations and transferring the nearest units, repeated until reconfiguration is complete.}
\label{fig:MARS-SHIFT}
\vspace{-7mm}
\end{figure*}

To bridge the gap, we propose TransforMARS, a general fault-tolerant self-reconfiguration framework that also plans disassemble and reassemble motions with controllability guarantees. Unlike prior methods, our approach handles multiple faulty units and generalizes to arbitrary configurations. See Figs.~\ref{fig:demo}(c) and (d) for examples. In particular,  Steps 1--2 of Fig.~\ref{fig:demo}(c) first relocate the normal units that may obstruct the trajectory of the faulty unit, which ensures that the faulty unit can be transferred to desired position without conflict in Step 3.
More specifically, our contributions are summarized as follows:
\begin{enumerate}
\item We propose a general self-reconfiguration planner for arbitrarily shaped MARS with multiple faulty units. For each faulty unit, the method \textbf{first} constructs a controllable subassembly that includes the faulty unit for transfer. Unlike prior approaches which build the MCS by selecting normal units directly from the original configuration without relocating them (black and blue dashed boxes in Fig.~\ref{fig:MARS-SHIFT}), our method flexibly transfers normal units that are not directly connected to the faulty unit and reassembles them into a MCS identified in the target configuration. \textbf{Second}, to ensure every MCS can reach its designated position, we identify and remove potential obstructions (“blocker units”) along their transfer trajectories (see the first two steps in Fig.~\ref{fig:demo}(c) and Step 2 in Fig.~\ref{fig:MARS-SHIFT}(c), Ours). This stage is called the path-clearance strategy. Once the trajectories are unobstructed, the MCS units are moved to their target positions. \textbf{Finally}, to prevent blockages caused by an incorrect assembly order, such as earlier units obstructing later ones, we compute a \textit{conflict-free assembly sequence}, ensuring that the remaining normal units are relocated to the vacant positions without interference. 

\item Extensive experiments show that our method can maintain the controllability of intermediate subassemblies and reconfigure with substantially fewer assembly and disassembly steps compared with ~\cite{SR-baseline2020}. In contrast to~\cite{SR-ICRA2025}, our algorithm extends self-reconfiguration from single-fault to multi-fault scenarios and supports arbitrary, irregular configurations. We further validate the approach on irregular shapes in CoppeliaSim \cite{coppeliaSim} simulation and execute the planned sequences on real quadrotors using Crazyswarm~\cite{crazyswarm}, demonstrating its feasibility.

\end{enumerate}

\section{Preliminaries}
We assume that all units in MARS are homogeneous, each with a unit mass $m$. Let $n$ denote the total number of drone units. We denote the overall control input of MARS as $\boldsymbol{u}_f=[T\ \tau_x\ \tau_y\ \tau_z]^\mathrm{T}$, where $T$ is the collective thrust generated by all units, and $\tau_x$, $\tau_y$, and $\tau_z$ are the collective torques along the $x$-, $y$-, and $z$-axes, respectively. The control input $\boldsymbol{u}_f$ is bounded by the maximum limit $\boldsymbol{u}_{\max}=[T_{\max}\ \tau_{x,\max}\ \tau_{y,\max}\ \tau_{z,\max}]^\mathrm{T}$ and the lower bound $\boldsymbol{u}_0=[0\ 0\ 0\ 0]^\mathrm{T}$. Accordingly, we define the feasible control input set as $\boldsymbol{\varOmega}=\{\boldsymbol{u}\mid \boldsymbol{u}_0 \leq \boldsymbol{u} \leq \boldsymbol{u}_{\max}\}$, which includes all possible control inputs that MARS can provide. When all units function properly, the system can hover in the air. However, under certain failure scenarios, specific configurations may become uncontrollable if the remaining normal units cannot compensate for the lost thrust and torque contributed by faulty units. To quantify controllability, we adopt the CM defined in~\cite{SR-ICRA2025}, whose formulation is given as follows. For set $\boldsymbol{\varOmega}$ and an arbitrary point $\boldsymbol{\alpha}$, we first define the following function $\zeta$ as an indicator:
\begin{equation}
    \hspace*{-0.23cm}
    \left. \mathrm{\zeta}\!\left( \boldsymbol{\alpha },\partial \boldsymbol{\varOmega } \right) \triangleq \left\{ \begin{array}{l}
	\underset{\boldsymbol{\beta }\in \partial \boldsymbol{\varOmega }}{\min} \!\left\| \boldsymbol{\alpha }\!-\!\boldsymbol{\beta } \right\|, \quad \text{if }\boldsymbol{\alpha }\!\in \boldsymbol{\varOmega }\\
	-\underset{\boldsymbol{\beta }\in \partial \boldsymbol{\varOmega }}{\min} \! \left\| \boldsymbol{\alpha }\!-\!\boldsymbol{\beta } \right\|, \quad \text{if } \boldsymbol{\alpha }\!\in \boldsymbol{\varOmega }^{\mathrm{C}} \\
\end{array} \right. \right. 
\label{eq:define CM}
\end{equation}
where $\partial \boldsymbol{\varOmega}$ denotes the boundary of the set $\boldsymbol{\varOmega}$, and $\boldsymbol{\varOmega}^C$ its complement. In essence, $\mathrm{\zeta}\!\left(\boldsymbol{\alpha},\partial \boldsymbol{\varOmega}\right)$ represents the signed minimum distance from the point $\boldsymbol{\alpha}$ to the boundary $\partial \boldsymbol{\varOmega}$, with the sign determined by the membership of $\boldsymbol{\alpha}$. Specifically, the distance is positive if $\boldsymbol{\alpha}$ lies inside $\boldsymbol{\varOmega}$, zero if it lies on the boundary, and non-positive otherwise.

For our problem, $\boldsymbol{\varOmega}$ is exactly the feasible control input set of MARS under failures. Let $\boldsymbol{g}=[nmg\ 0\ 0\ 0]^\mathrm{T}$ (with $g$ being the gravitational acceleration). If $\boldsymbol{g}$ is an \textit{interior point} of $\boldsymbol{\varOmega}$, then the CM takes a positive value given by
\begin{equation}
\zeta\!\left(\boldsymbol{g}, \partial\boldsymbol{\varOmega}\right)
= \min_{\boldsymbol{u}_f \in \partial \boldsymbol{\varOmega}}
\left\| \boldsymbol{g} - \boldsymbol{u}_f \right\| > 0,
\end{equation}
which indicates that the available control authority is sufficient to counteract gravity and stabilize the orientation, allowing MARS to continue hovering even after a failure. In contrast, when $\zeta\!\left(\boldsymbol{g},\partial\boldsymbol{\varOmega}\right)<0$, the system becomes uncontrollable. Moreover, a larger value of $\zeta\!\left(\boldsymbol{g},\partial\boldsymbol{\varOmega}\right)$ corresponds to greater control authority, providing a straightforward and effective metric for evaluating the control performance of the system. Previous work \cite{SR-ICRA2025} shows that MARS configurations with higher CM values achieve better tracking performance; therefore, we use CM as the metric for self-reconfiguration.

\section{Generalized Self-Reconfiguration of MARS}
In this section, we present a generalized self-reconfiguration method applicable to arbitrarily shaped configurations with multiple faulty units. First, find the optimal configuration. Second, \textbf{\Cref{algorithm1}} determines the MCS for each faulty unit, ensuring that faulty units are not isolated during the reconfiguration process. Third, \textbf{\Cref{algorithm2}} plans path clearance and transfer for the Virtual Minimum Controllable Subassembly (VMCS), along with a conflict-free assembly sequence that transfers all faulty units to the target configuration while preventing blockage during the reconfiguration process. Finally, transfer normal units to optimize the distance.
\subsection{Find Optimal Configuration}
Let $p_i = (x_i, y_i)$ denote the coordinate of the $i$-th unit in the MARS $\mathcal{P}$, 
and let $n_f$ be the number of faulty units, with $\mathcal{F}$ denoting the set of faulty units. During the self-reconfiguration process, the single assembled MARS can be decomposed into $n_p$ subassemblies. We denote their collection by
\begin{subequations} \begin{align} &\mathcal{P} = \{p_1, p_2, \dots, p_n\}, \\[6pt] 
&\pi_0\!\big(G[\mathcal{P}]\big) = \{\mathcal{S}_1, \mathcal{S}_2, \cdots, \mathcal{S}_{n_p}\}, \\[6pt] 
&\mathcal{S}_i=\bigcup_j\{p_{i,j}\},
i\in\{1,2,\cdots,n_p\}, \end{align} \end{subequations}
where, $G[\mathcal{P}]$ denotes the grid graph induced by the set of occupied cells $\mathcal{P}$, where edges represent adjacency (i.e., 4-neighbor connectivity). 
The operator $\pi_0(\cdot)$ returns the set of connected components of a graph. 
Thus, $\pi_0(G[\mathcal{P}]) = \{\mathcal{S}_1, \cdots, \mathcal{S}_{n_p}\}$ is the partition of $\mathcal{P}$ into $n_p$ disjoint subassemblies, each $\mathcal{S}_i$ consisting of units $p_{i,j}$. 
We denote by $\mathcal{D}_i$ the subset of faulty units in $\mathcal{S}_i$. 
Accordingly, a MARS as a whole or any of its subassemblies can be fully described by the pair $(\mathcal{P},\mathcal{F})$ or $(\mathcal{S}_i,\mathcal{D}_i)$, respectively.

For subassemblies that contain at least one faulty unit, their CMs are computed using~\eqref{eq:define CM} and the smallest value among them is employed to measure the system-level CM. Specifically, suppose $n_d$ subassemblies contain faulty units; the overall CM of MARS is defined as
\begin{equation}
CM_{(\mathcal{P},\mathcal{F})} = \min_{i=1,2,\cdots,n_d} \zeta\!\left(\mathbf{g}_i, \partial\boldsymbol{\varOmega}_i\right),
\end{equation}
where $\boldsymbol{\varOmega}_i$ is the feasible control input set of the $i$-th faulty subassembly, $\mathbf{g}_i$ is its total weight. The indicator $\zeta\!\left(\mathbf{g}_i, \partial\boldsymbol{\varOmega}_i\right)$ denotes its CM value as computed in~\eqref{eq:define CM}. In the following sections, we use $CM_{(\mathcal{P},\mathcal{F})}$ to evaluate MARS during reconfigurations. 

We determine the optimal configuration $(\mathcal{P},\mathcal{F})$, defined as the arrangement that maximizes CM among all configurations with the same number of faulty and normal units (Fig.~\ref{fig:MARS-SHIFT}(a)). 
\subsection{Find Virtual Minimum Controllable Assembly}
Prior work~\cite{SR-baseline2020} transfers faulty units by attaching only a single functional unit. When four rotors fail, the resulting configuration yields a negative CM and becomes practically infeasible (Fig.~\ref{fig:MARS-SHIFT}(c), \textbf{Ref.~[9]}). In general, each faulty unit must be bolstered by multiple normal units to form a flyable subassembly, termed the MCS~\cite{SR-ICRA2025} (Fig.~\ref{fig:MARS-SHIFT}(c)-\textbf{Ref. [10]}). However,~\cite{SR-ICRA2025} identifies the MCS only from the units already attached to the faulty unit, which is conservative.
In contrast, we consider all available normal units in the MARS when constructing the MCS, including those not directly attached to the faulty unit (see Fig.~\ref{fig:MARS-SHIFT}(c)-\textbf{Ours}, for example).
We refer to this more flexible and general construction as the VMCS. The detailed procedure is presented in~\Cref{algorithm1}, and an example calculation is illustrated in Fig.~\ref{fig:MARS-SHIFT}(b).

Specifically, if a faulty unit is already connected with sufficient normal units to form a VMCS ($\mathcal{P}^\dagger, \mathcal{F}^\dagger$), no additional disassembly or assembly is required. Otherwise, when a VMCS cannot be found for a faulty unit in the original configuration, i.e., vacant position $p^*$ exists, we detach a normal unit $p_{i,j}$ and reassemble it with the faulty one, as shown in Fig.~\ref{fig:MARS-SHIFT}(b) and in the Fig.~\ref{fig:MARS-SHIFT}(c)-\textbf{Ours}. To identify the optimal normal unit to detach, we formulate the following disassembly optimization problem that jointly considers CM and path length. The objective is to select a unit whose removal results in a safer remaining MARS configuration (higher CM) while minimizing travel distance to avoid detours:
\begin{subequations}
    \begin{align}
         \min_{p_{i,j}\notin\mathcal{F}} &\;  c_1\Delta _{CM}^{2} 
- c_2 L(p_{i,j}, p^*) \label{eq:optimization},\\
    & \Delta_{CM}=CM_{(\mathcal{P}_{\mathcal{X}},  \mathcal{F}_{\mathcal{X}})} - CM_{(\mathcal{P}^*, \mathcal{F}^*)},
    \end{align}
\end{subequations}
where $(\mathcal{P}_{\mathcal{X}},  \mathcal{F}_{\mathcal{X}})$ denotes the MARS configuration after detaching unit $p_{i,j}$, $c_1$ and $c_2$ are weight parameters balancing the CM deviation and the path length, respectively. Here, $L(p_{i,j}, p^*)$ denotes the path length computed using the $A^*$ algorithm for moving the $j$-th unit in the $i$-th subassembly from its current position $p_{i,j}$ to the destination $p^*$.
 
\subsection{Faulty Units Transfer: Path Clearance and Transfer for VMCS}
Some normal units may obstruct the transfer of VMCS, as illustrated in the Fig.~\ref{fig:MARS-SHIFT}(c)-\textbf{Ours}. In such cases, the first two steps relocate these normal units to alternative positions, enabling the VMCS to reach its designated destination. We define such interfering units as \emph{blocker units} if they (i) occupy a target position assigned to a VMCS or (ii) lie along a VMCS's transfer trajectory. To ensure successful VMCS transfer, these blocker units must temporarily vacate their current positions.
\begin{figure}[t]
\vspace{-8pt}
\begin{algorithm}[H]
\scriptsize
\caption{VMCS Identification Algorithm}\label{algorithm1}
\begin{algorithmic}[1]
\State \textbf{Input:} $\mathcal{P}, \mathcal{F},k=1,CM^*=-\inf$
\State \textbf{Output:} VMCS $\mathcal{P}^\dagger, \mathcal{F}^\dagger$
    \While{ $CM^*<0$}
        \State $\mathbf{C}_k \leftarrow$ the set of all possible rigidly connected assemblies $\{\mathcal{P}', \mathcal{F}'\}$ with $k$ normal units and the faulty unit.
        \State $CM^* \leftarrow\max \{CM_{(\mathcal{P}', \mathcal{F}')}$ for $\forall (\mathcal{P}', \mathcal{F}')\in\mathbf{C}_k\}$
        \State $k\leftarrow k+1$
    \EndWhile
        \State \textbf{Return} $\mathcal{P}^\dagger, \mathcal{F}^\dagger$
\end{algorithmic}
\end{algorithm}
\vspace{-12mm}
\end{figure}
Instead of relocating blocker units to arbitrary or manually specified positions as in~\cite{SR-baseline2020}, we move them to a set of temporary \emph{waiting positions} $\mathcal{W} = \{w_1, w_2, \dots, w_q\}$,
selected from the vacant positions in the target configuration $\mathcal{P}^*$ that are not part of any VMCS transfer path. Formally,
$ \mathcal{W} =
\bigl\{ w \mid
w \in \mathcal{P}^* \setminus
\bigl( (\mathcal{P} \cap \mathcal{P}^*) \cup T \bigr)
\bigr\}$,
where $T$ denotes the set of transfer path coordinates.  
This strategy, which is termed the \emph{path-clearance relocation rule}, reduces redundant moves by ensuring that vacated units remain close to their final target positions. As detailed in \textbf{\Cref{algorithm2}}, Lines~14--25, each blocker unit is first moved to its nearest waiting position $w^* \in \mathcal{W}$ (Line~21). Once these paths are cleared, the VMCS units proceed to their designated targets.

\begin{figure}[t]
\vspace{-8pt}
\begin{algorithm}[H]
\scriptsize
\caption{TransforMARS Algorithm}\label{algorithm2}
\begin{algorithmic}[1]
\State \textbf{Input:} $\mathcal{P}, \mathcal{F}$
\State \textbf{Output}: $\mathcal{P},\mathcal{F}$
\State  $\mathcal{P}^*, \mathcal{F}^*$ $\leftarrow$ Compute optimal configuration for $(\mathcal{P}, \mathcal{F})$~\cite{SR-ICRA2025}
\State $\mathcal{P}^\dagger, \mathcal{F}^\dagger \leftarrow$ Compute VMCS for all faulty units in $\mathcal{F}$
\For{$p_f\in\mathcal{F}$}\Comment{Find VMCS}
    \If{cannot find adjacent normal units to form VMCS}
        \For{all vacant position $p^*$ in VMCS}
            \State $p_{i,j}=$ optimization$(\mathcal{P}, \mathcal{F}, \mathcal{P}^*,\mathcal{F}^*,p^*)$~\eqref{eq:optimization}
            \State transfer $p_{i,j}$ to $p^*$ using trajectory generated by $A^*$
            \State update $\mathcal{P},\mathcal{F}$
        \EndFor
    \EndIf
\EndFor

\For{$i$-th VMCS}\Comment{Path clearance for VMCS}
    \State Compute trajectory $T_i$ of transfering $i$-th VMCS s.t. faulty unit reaching position $p_f\in\mathcal{F}^*$  
\EndFor
\State $T\leftarrow T_1\cup T_2\cup\cdots\cup T_{n_f}$
\For{blocker unit $j$ staying in trajectory $T$}
    \State waiting position set $\mathcal{W}=\{w \mid w \in \mathcal{P}^*\setminus\big(\mathcal{ P}\cap\mathcal{P}^*\cup T \big)\}$
    \State $w^*=\arg\min_{w\in \mathcal{W}} L(w,p_j)$
    \State transfer $p_j$ to $w^*$ using trajectory generated by $A^*$
    \State update $\mathcal{P},\mathcal{F}$
\EndFor
    \State transfer all VMCS s.t. faulty unit reaching position $p_f\in\mathcal{F}^*$ 
    \State update $\mathcal{P},\mathcal{F}$
\While{$\mathcal{P}^* \setminus (\mathcal{P} \cap \mathcal{P}^*)\neq \varnothing$}\Comment{Conflict-free destination}
    \State compute virtual start point $p_s$
    \State $\mathcal{P}_t=\mathcal{P}^* \setminus (\mathcal{P} \cap \mathcal{P}^*)$ 
    \For{$p_t\in\mathcal{P}_t$}
        \State $T\leftarrow$trajectory coordinates set from $p_s$ to $p_t$ using $A^*$ 
        \While{$\exists p'\in\mathcal{P}_t\cap T$ and $p'\neq p_t$}
            \State $\mathcal{P}_t=\mathcal{P}_t\setminus\{p'\}$
        \EndWhile
    \EndFor
    \State transfer units with shortest path to $\mathcal{P}_t$ using \eqref{eq:binary}
    \State update $\mathcal{P},\mathcal{F}$
\EndWhile
\State \textbf{Return} $\mathcal{P},\mathcal{F}$
\end{algorithmic}
\end{algorithm}
\vspace{-10mm}
\end{figure}
\subsection{Normal Units Transfer: Conflict-Free Assembly Sequence}

Previous work~\cite{SR-ICRA2025} plans the disassembly and assembly sequence by always selecting the option that maximizes the CM at each step, but its applicability is limited to standard rectangular configurations. In fact, once the CM exceeds a fixed threshold, the control authority of the entire system is sufficient for the designed controller. We introduce a more flexible strategy that can be applied to arbitrary configurations. This process is repeated until $\mathcal{P} = \mathcal{P}^*$.

\subsubsection{Conflict-Free Destination for Next Unit} 
During reassembling the remaining normal units, we initialize their corresponding set of \emph{target positions} as $\mathcal{P}_{t}:=\mathcal{P}^* \setminus (\mathcal{P} \cap \mathcal{P}^*)$. To achieve efficient re-assembly, we need to identify an assembly sequence that guarantees accessibility, meaning that no earlier-placed unit obstructs the movement of subsequent units.

Specifically, for each step, a virtual start point $p_s$ is chosen near $\mathcal{P}^*$ but outside $\mathcal{P} \cup \mathcal{P}^*$. An $A^*$ path is then computed from $p_s$ to each candidate target position. If a candidate target lies along the $A^*$ path to another target, as illustrated in the conflict-free stage of Fig.~\ref{fig:MARS-SHIFT}(d) (Units 2--8 on the orange path and Unit 10 on the yellow path), it is discarded to avoid blockage. The remaining target positions in the set $\mathcal{P}_t$ (see \textbf{\Cref{algorithm2}}, Line 26--33) are thus guaranteed not to obstruct the paths of subsequent transfers.

\subsubsection{Reconfiguration Planner}
After determining the destination in $\mathcal{P}_t$, we next identify the unit to be relocated. For each candidate unit in set $\mathcal{P}_c:=\mathcal{P}\setminus(\mathcal{P}\cap\mathcal{P}^*)$, we compute the $A^*$ path length from its current position to the destination. The unit with the minimum path length is then selected and transferred to the destination. 

When $|\mathcal{P}_t|>1$, the normal units may be transferred to different destinations. To assign normal units $p_{c,j} \in \mathcal{P}_c$ to target positions $p_{t,i} \in \mathcal{P}_t$ such that the total transfer distance is minimized, we formulate a binary assignment optimization problem:

\begin{equation}
\begin{aligned}
\min_{x_{ij}} \quad & \sum_{i=1}^{|\mathcal{P}_t|} \sum_{j=1}^{|\mathcal{P}_c|} x_{ij} \cdot L(p_{c,j}, p_{t,i}) \\
\text{s.t.} \quad 
& \sum_{j=1}^{|\mathcal{P}_c|} x_{ij} = 1, \quad \forall i = 1,\dots,|\mathcal{P}_t| \quad  \\
& \sum_{i=1}^{|\mathcal{P}_t|} x_{ij} \leq 1, \quad \forall j = 1,\dots,|\mathcal{P}_c| \quad  \\
& x_{ij} \in \{0,1\}
\end{aligned}
\label{eq:binary}
\end{equation}
Here, $x_{ij}$ is a binary variable indicating whether candidate unit $p_{c,j}$ is assigned to target position $p_{t,i}$, constraint $\sum_{j=1}^{|\mathcal{P}_c|} x_{ij} = 1$ indicates that each target $p_{t,i}$ is assigned exactly one unit, constraint $\sum_{i=1}^{|\mathcal{P}_t|} x_{ij} \leq 1$ indicates that each normal unit $p_{c,j}$ is scheduled at most once, and $L(p_{c,j}, p_{t,i})$ denotes the transfer distance between them.

\section{Evaluation and Real-World Experiments}
We employ a high-fidelity quadrotor model in CoppeliaSim~\cite{coppeliaSim} following the setup in~\cite{SR-ICRA2025}, with a controller adopted from~\cite{HUANG2024416,huang2025robust-iros} that supports both symmetric and asymmetric configurations. The experiments evaluate the controllability of arbitrarily shaped MARS with multiple faulty units that cannot be addressed by~\cite{SR-baseline2020,SR-ICRA2025}, analyzing the CM as well as disassembly and assembly steps under failures. In particular, we compare the proposed method against~\cite{SR-baseline2020,SR-ICRA2025} on standard $M\times N$ configurations.

\subsection{Flexible Self-Reconfiguration}
\subsubsection{Standard M $\times$ N Configuration with Multiple Faulty Units}
\begin{figure}[!t]
\centering
\includegraphics[width=3.4in]{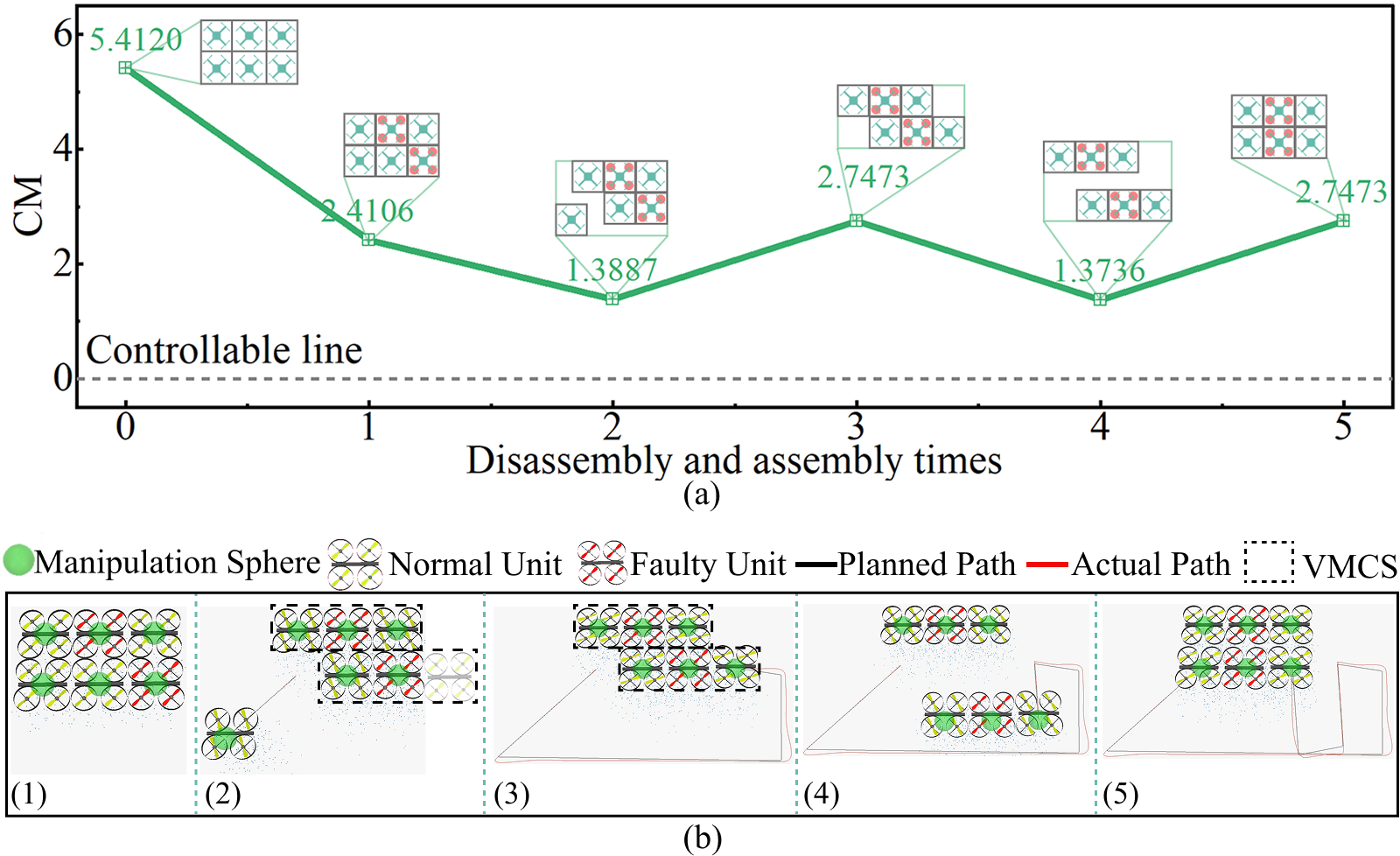}
\vspace{-5 mm}
\caption{Self-reconfiguration process of a $3\times2$ assembly. (a) Controllability margin versus disassembly and assembly step with corresponding configurations shown. (b) Dynamical simulation of the self-reconfiguration sequence in CoppeliaSim.}
\label{fig:M*N 2+6 simulation}
\vspace{-4 mm}
\end{figure}

Unlike previous work~\cite{SR-ICRA2025} that considers only single-fault cases in the $3 \times 2$ configuration, we evaluate this configuration with two faulty units. Fig.\ref{fig:M*N 2+6 simulation}(a) illustrates the reconfiguration sequence and corresponding CM analysis. The algorithm completes the sequence in four disassembly and assembly steps (2--5), achieving a minimum CM of 1.3736, which is sufficient for stable control. The dynamic reconfiguration simulation results, obtained using the fault-tolerant controller in\cite{HUANG2024416, huang2025robust-iros}, are shown in Fig.\ref{fig:M*N 2+6 simulation}(b). In Step 2, after identifying the VMCS with \textbf{\Cref{algorithm1}}, normal units detach and relocate as shown in Step 2--3, ensuring that all faulty units can be transferred via VMCS. Subsequently, Algorithm~\ref{algorithm2} transfers the two VMCS to their final positions in the optimal configuration (Steps~4--5).

\begin{figure}[!t]
\centering
\includegraphics[width=3.4in]{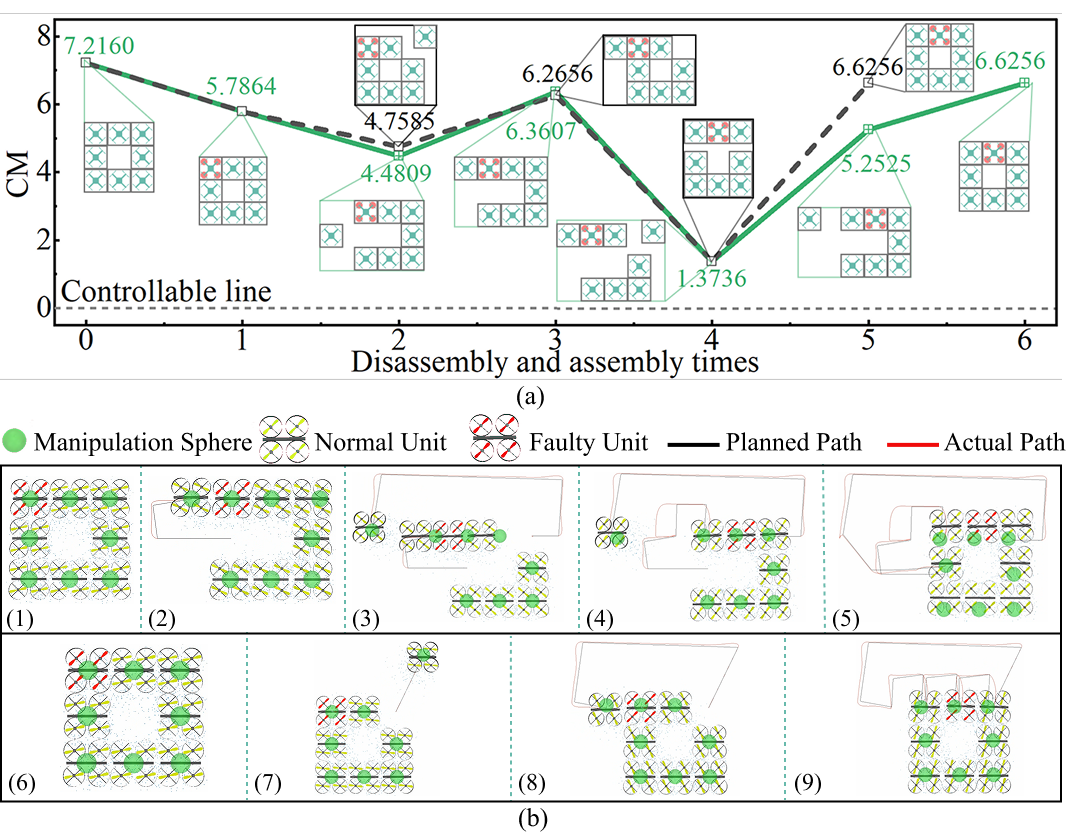}
\vspace{-6 mm}
\caption{Self-reconfiguration process of a $3\times3$ hollow assembly. (a) CM versus disassembly and assembly times. A higher $c_1$ weight (black dashed line) yields a higher average CM than the lower $c_1$ setting (green line). The black dashed line also results in a shorter reconfiguration sequence, as path clearance is skipped (Step 4 in green line) due to the relocation of normal unit (No.~3, top right). (b) Dynamic simulation in CoppeliaSim: (1)--(5) show the sequence for the green line, and (6)--(9) for the black dashed line. }
\label{fig:hollow simulation}
\vspace{-5mm}
\end{figure}

\begin{figure*}[!t]
\centering
\includegraphics[width=6.9in]{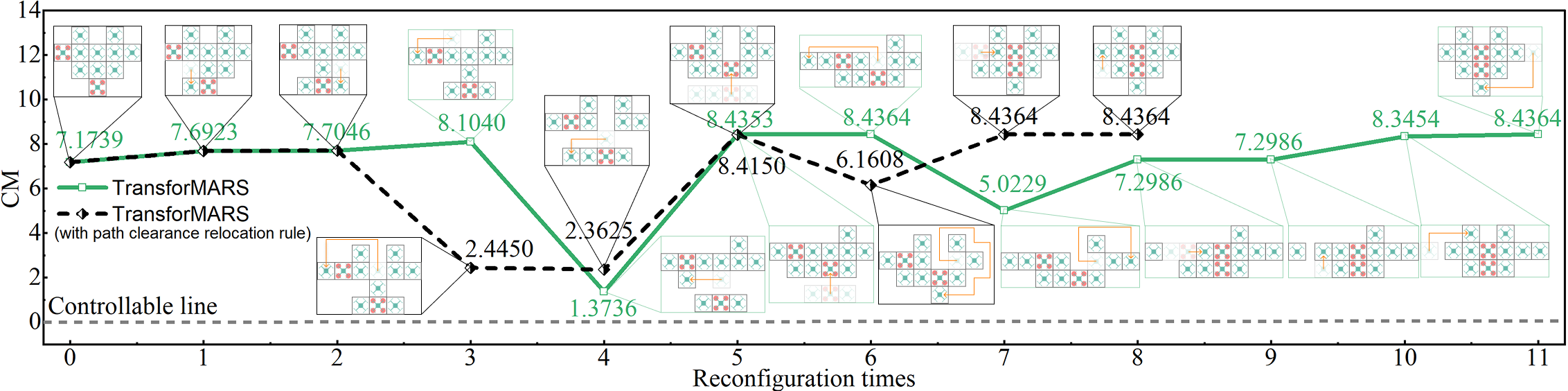}
\vspace{-3mm}
\caption{Self-reconfiguration sequence of an 11-unit heart-shaped assembly. The black dashed line and green line correspond to the reconfiguration methods with and without the path clearance relocation rule, respectively. With the relocation rule, the number of reconfiguration steps is reduced, demonstrating its effectiveness in improving self-reconfiguration efficiency.
}
\label{fig:heart-shaped simulation}
\vspace{-5mm}
\end{figure*}

\subsubsection{Hollow Configuration}
We further evaluate a hollow configuration. Fig.~\ref{fig:hollow simulation}(a) illustrates two self-reconfiguration sequences of a $3\times3$ hollow configuration with one faulty unit (No.~1, top left, marked in red) under different weight parameter settings, while Fig.~\ref{fig:hollow simulation}(b) shows the corresponding simulations. Specifically, we set $(c_1,c_2)=(2,-0.1)$ for the green line and $(c_1,c_2)=(4,-0.1)$ for the black dashed line in \eqref{eq:optimization}. For small-scale configurations, assigning a higher weight to the CM term increases the average CM value, whereas the path-length weight has little impact due to short paths at each step. Thus, to ensure controllability, prioritizing the CM weight is essential. Notably, the disassembly and assembly times of the black dashed sequence are fewer than those of the green one, since unit (No.~3, top right) serves as the best candidate to form a VMCS, thereby avoiding extra disassembly and assembly steps (see black dashed line, Step 3) in path clearance.
\begin{figure}[!t]
\centering
\includegraphics[width=3.4in]{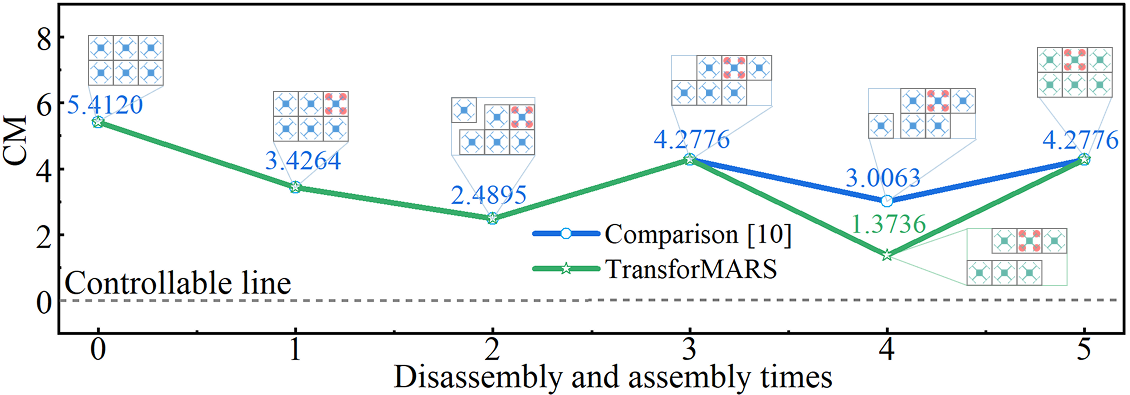}
\vspace{-8mm}
\caption{Comparison of disassembly and assembly steps and CM for a failure at position 3 in a $3\times2$ assembly. TransforMARS (green) requires no extra steps compared with~\cite{SR-ICRA2025} (blue). Although the CM decreases at step 4, the system remains controllable.
 }
\label{fig:Comparison in a 3x2 assembly}
\vspace{-5mm}
\end{figure}

\subsubsection{Arbitrarily Shaped Configuration with Multiple Faulty Units}
Unlike~\cite{SR-ICRA2025}, which addresses only standard rectangular configurations, we further evaluate arbitrarily shaped assemblies, such as an 11-unit heart-shaped configuration in Fig.~\ref{fig:heart-shaped simulation}. The green line denotes transferring units to a waiting position within the same row, while the black dashed line applies the \textit{path clearance relocation rule} to move units directly to vacant coordinates in the optimal configuration. With this strategy, the reconfiguration sequence is reduced by three steps (from 11 to 8), equivalent to six disassembly-assembly operations, while also avoiding long detours: without it, the total path length increases by $35.29\%$ (from 34 to 46 unit distances). These results highlight the effectiveness of the path clearance relocation rule in reducing both path length (energy consumption) and reconfiguration times, thereby improving efficiency and safety.

\begin{figure*}[!t]
\centering
\includegraphics[width=6.9in]{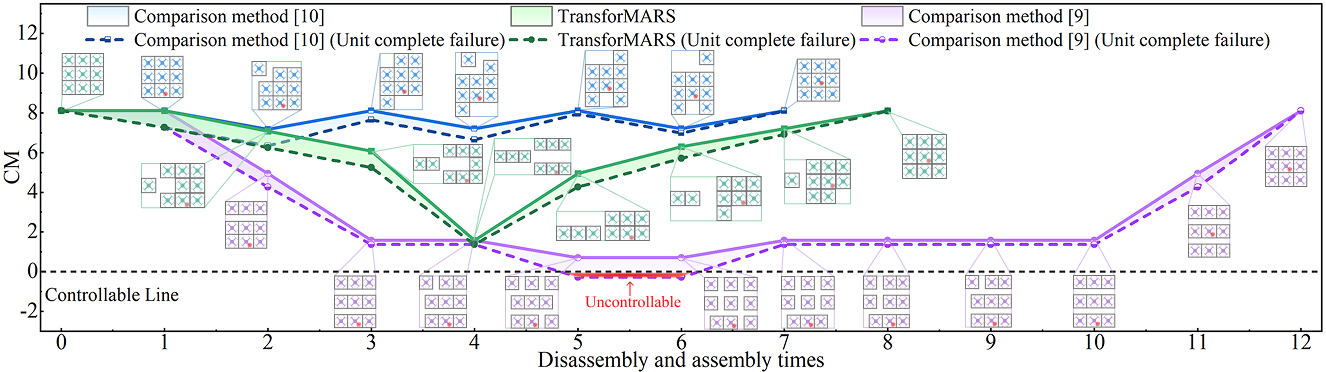}
\vspace{-3mm}
\caption{Comparison of disassembly and assembly times and CM for failure at position 8 in a 3$\times$3 assembly. The solid line represents single rotor failure and dashed line represents unit complete failure. Compared with \cite{SR-baseline2020}, TransforMARS achieves substantially higher CM with fewer steps. Relative to \cite{SR-ICRA2025} (blue), the average CM decreases by only 20.50\%, and the number of steps increases by only 16.67\%.}
\label{fig:Comparison in a 3x3 assembly}
\vspace{-3mm}
\end{figure*}

\subsection{Comparison with the Baseline Method}
Table~\ref{tab:comparison} compares existing reconfiguration methods~\cite{SR-baseline2020, SR-ICRA2025} with the proposed TransforMARS. \cite{SR-baseline2020} addresses rotor-level failures and can handle multiple rotor faults in arbitrary configurations, but it overlooks the controllability of intermediate subassemblies by simply connecting a faulty unit to a single normal one. \cite{SR-ICRA2025} extends reconfiguration to both rotor- and unit-level failures with explicit controllability analysis, yet it is restricted to single-fault cases in standard rectangular configurations. In contrast, the proposed TransforMARS supports multiple faults at both rotor and unit levels, applies to arbitrary configurations, and maintains subassembly controllability, thereby overcoming the limitations of prior approaches. In the following subsection, we compare the reconfiguration sequences of the baseline methods with those of TransforMARS, and calculate the CM value of MARS in each step.

\subsubsection{$3 \times 2$ Configuration}
We first compare our method with previous approach~\cite{SR-ICRA2025} on standard configurations. As shown in Fig.~\ref{fig:Comparison in a 3x2 assembly}, the self-reconfiguration sequences generated by~\cite{SR-ICRA2025} (blue line) and our proposed TransforMARS (green line) are illustrated. Compared with~\cite{SR-ICRA2025}, our method achieves the target configuration (Step 5) with the same number of disassembly and assembly steps, while the average CM decreases by only 11.52\% due to the transfer of VMCS. This demonstrates that, unlike~\cite{SR-ICRA2025}, which is restricted to limited configurations, TransforMARS not only solves standard $M\times N$ configuration problems with minor CM loss but also generalizes to a broader range of scenarios, including arbitrary shapes such as hollow and heart configurations.

\subsubsection{$3 \times 3$ Configuration}
To further demonstrate the robustness of our method, we compare the CM values at the intermediate stages of different approaches in a 3$\times$3 self-reconfiguration scenario with one faulty unit. In Fig.~\ref{fig:Comparison in a 3x3 assembly}, the self-reconfiguration sequences generated by different approaches are plotted, along with the line charts representing their intermediate CM values. Solid lines present scenarios with a single rotor failure, while dashed lines denote complete unit failure cases. For CM, our proposed method (green) achieves higher values, with an average improvement of up to 78.03\% compared to~\cite{SR-baseline2020} (purple). The average CM decreases by only 20.50\% relative to~\cite{SR-ICRA2025} (blue). In terms of disassembly and assembly, our approach requires 16.67\% more steps than \cite{SR-ICRA2025}, yet still reduces the number of steps by 36.36\% compared to~\cite{SR-baseline2020}. 

However, for both $ 3\times 2$ and $3 \times3$ configurations with one faulty unit, previous work ~\cite{SR-ICRA2025} mainly focuses on CM and exhibits limited generalization, as it only addresses the standard $M\times N$ case. In fact, as long as $\mathrm{CM} > 0$, the MARS remains controllable. For example, in Step~4 of Fig.~\ref{fig:Comparison in a 3x3 assembly}, the CM reaches $1.3736$, which is sufficient for stable control. In contrast, our approach extends to a wide range of configurations (e.g., hollow structures in Fig.~\ref{fig:hollow simulation}, arbitrary assemblies in Fig.~\ref{fig:heart-shaped simulation}, and multi-error units) and further incorporates accessible path planning for unit movement, as demonstrated in the previous section. This offers greater practical significance than methods that only emphasize CM.
\begin{table}[t]
\caption{Comparison of previous methods}
\label{tab:comparison}
\vspace{-2mm}
\centering
\scriptsize
\renewcommand{\arraystretch}{1.3}
\begin{tabular}{@{}lcccccc@{}}
\toprule
\makecell {\textbf{Method}} 
& \makecell{\textbf{Fault}} 
&  \makecell{\textbf{\# Faulty Units}} 
&  \makecell{\textbf{Configuration}} 
&  \makecell{\textbf{Safety} \\ \textbf{Guarantee}} \\
\midrule
Ref.~\cite{SR-baseline2020} & rotor & multiple rotors & \textbf{arbitrary} & \ding{55} \\
Ref.~\cite{SR-ICRA2025} & \textbf{rotor \& unit} & \makecell{single} & rectangular & \checkmark \\
\textbf{Ours} & \textbf{rotor \& unit} & \makecell{\textbf{multiple}} & \textbf{arbitrary} & \checkmark \\
\bottomrule
\end{tabular}
\vspace{-8mm}
\end{table}
\begin{figure*}[!t]
\centering
\includegraphics[width=6.9in]{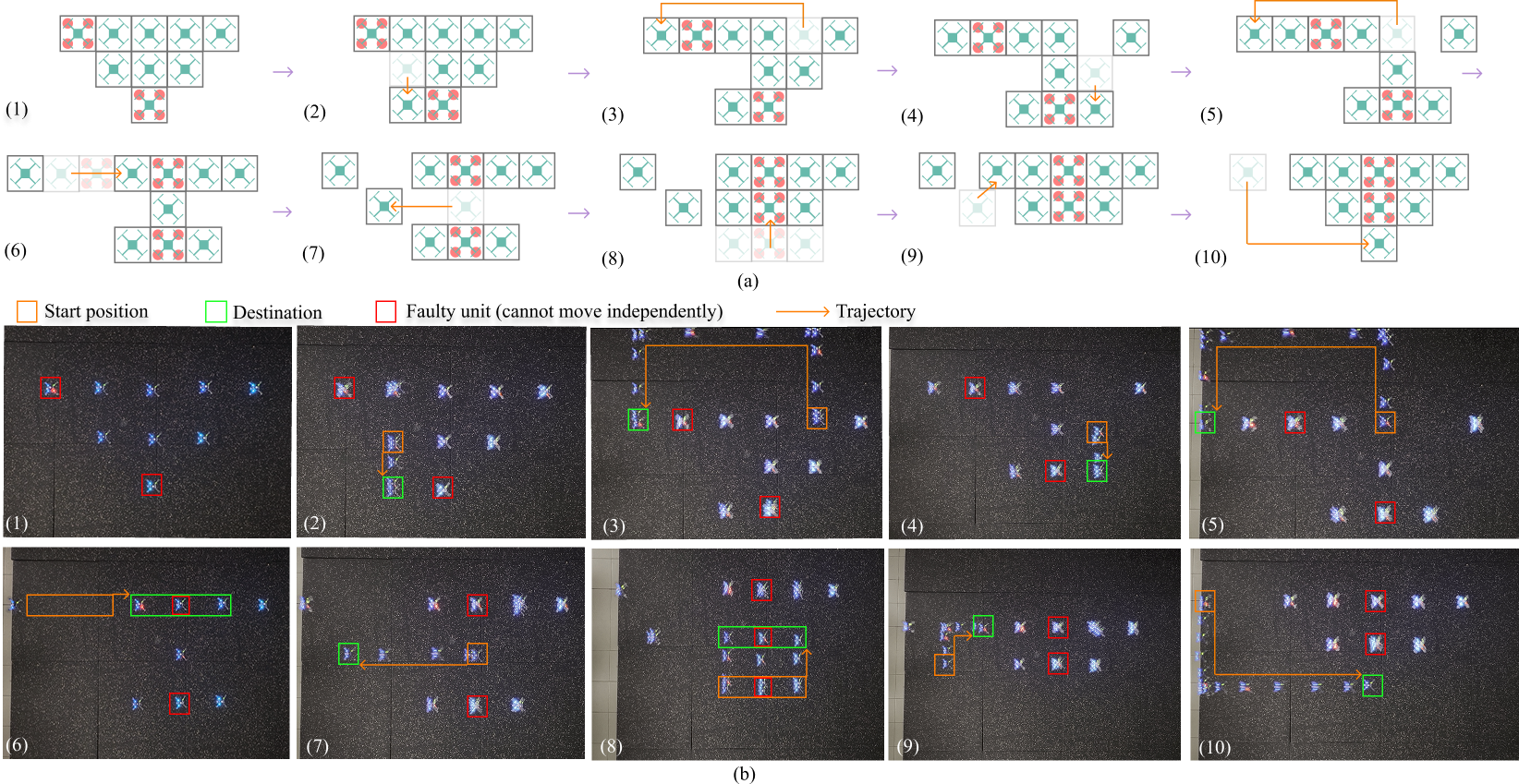}
\vspace{-4 mm}
\caption{ Real-world validation of self-reconfiguration.
(a) Reconfiguration sequence from the triangle configuration with two faulty units. (b) Experiments with 9 Crazyflie drones showing feasible trajectories during the self-reconfiguration process. A video and additional demonstrations are included in the supplementary video submitted with this paper.}
\label{fig:real-world}
\vspace{-4mm}
\end{figure*}
\subsection{Real-World Experiment}
We conduct real-world experiments to validate the proposed TransforMARS, a self-reconfiguration framework with integrated path planning. The testbed consists of nine Crazyflie micro-quadrotors serving as MARS units. A faulty unit cannot move independently and is transported by the VMCS determined by our planner. In contrast to previous approaches that only considered the reconfiguration sequence, where normal units were placed at manually designed waiting positions \cite{SR-baseline2020} or followed pre-defined trajectories \cite{SR-ICRA2025}, our method jointly plans both the accessible self-reconfiguration sequence and feasible trajectories for arbitrary configurations. Fig.~\ref{fig:real-world}(a) illustrates the reconfiguration sequence and trajectories from a triangle configuration with two faulty units, and Fig.~\ref{fig:real-world}(b) shows real-world experiments where Crazyflie drones follow planned trajectories to achieve successful self-reconfiguration. 

Specifically, Steps 2--4 in Fig.~\ref{fig:real-world} illustrate the construction of the VMCS. Steps 6 and 8 show the transfer of the VMCS to its final position in the target configuration, while Steps 5 and 7 demonstrate path clearance for the VMCS. Finally, Steps 9 and 10 depict the accessible disassembly and assembly sequences in which normal units complete the target configuration.

It is important to emphasize that our experiments focus exclusively on the trajectory execution and coordination phase of self-reconfiguration. In comparison, full self-reconfiguration, which involves both physical docking and separation, presents greater challenges due to the requirement of continuous and repeated operations. These operations remain inherently unstable, as existing docking methods~\cite{Modquad, ModQuad-Vi} and separation mechanisms~\cite{separation-2019design} already exhibit noticeable instability after just a single docking or separation cycle. Moreover, full self-reconfiguration must repeatedly perform such unstable procedures until the system reaches its target configuration, making the process considerably more complex and error-prone than executing a single docking or separation action.

\section{Conclusion and Future Work}
To address rotor- and unit-level failures in MARS, this paper presents a flexible self-reconfiguration framework that enhances fault-tolerant control. The proposed approach integrates self-reconfiguration with path planning for transporting faulty units, and supports reconfiguration into arbitrary target configurations. Specifically, our algorithms (i) identify the VMCS to enable safe transfer of faulty units, (ii) perform path clearance to guide the VMCS to their designated positions, and (iii) plan accessible disassembly and assembly sequences to avoid blockage at each stage. Compared with existing approaches, this work demonstrates the first general self-reconfiguration method that handles multiple faulty units in arbitrary configurations. For future work, the existing docking~\cite{Modquad, ModQuad-Vi} and separation~\cite{separation-2019design} mechanisms of MARS still face several challenges that can significantly impact self-reconfiguration performance. Until now, no successful real-world self-reconfiguration experiments have been reported. Therefore, we plan to conduct real-world experiments incorporating both disassembly and assembly functions to further validate the proposed framework.

\bibliographystyle{IEEEtran}

\bibliography{reference}

\end{document}